\documentclass[runningheads]{llncs}
\pdfoutput=1
\usepackage{graphicx}
\usepackage{booktabs} 
\usepackage[utf8]{inputenc}
\usepackage{CJKutf8}
\usepackage{makecell}
\usepackage{caption}
\usepackage{amssymb}
\usepackage{amsmath}
\usepackage{multirow}
\usepackage{bbding}
\usepackage{colortbl}
\usepackage{xcolor}
\usepackage{CJK}
\usepackage{url}
\usepackage{comment} 
\usepackage{adjustbox} 
\usepackage{wrapfig}

\usepackage{marvosym}
\usepackage[
bookmarks=true,
colorlinks=true,
linkcolor=red,
urlcolor=red,
citecolor=blue,
bookmarks=true,
linktocpage=true,   
hyperindex=true
]{hyperref}
\hypersetup{
    colorlinks=true,
    linkcolor=red,    
    citecolor=blue,    
    urlcolor=red,     
    breaklinks=true,
    pdfborder={0 0 0}  
}

\begin{document}

\title{MFH: Marrying Frequency Domain\\ with Handwritten Mathematical\\ Expression Recognition}
\author{Huanxin Yang$^{1\star}$\textsuperscript{(\Letter)} 
        \and Qiwen Wang$^{1}$\thanks{Equal contribution.}\textsuperscript{(\Letter)}}
\institute{$^1$School of Future Technology, \\
Huazhong University of Science and Technology, 
Wuhan 430074, China \\
\email{\{hxyang,qwwang\}@hust.edu.cn}}

\maketitle 
\begin{abstract}
Handwritten mathematical expression recognition (HMER) suffers from complex formula structures and character layouts in sequence prediction. In this paper, we incorporate frequency domain analysis into HMER and propose a method that \textbf{m}arries \textbf{f}requency domain with \textbf{H}MER (MFH), leveraging the discrete cosine transform (DCT). We emphasize the structural analysis assistance of frequency information for recognizing mathematical formulas. When implemented on various baseline models, our network exhibits a consistent performance enhancement, demonstrating the efficacy of frequency domain information. Experiments show that our MFH-CoMER achieves noteworthy accuracy rates of 61.66\%/62.07\%/63.72\% on the CROHME 2014/2016/2019 test sets. The source code is available at \url{https://github.com/Hryxyhe/MFH}.
\keywords{Handwritten mathematical expression recognition \and Frequency domain analysis \and Discrete cosine transform.}
\end{abstract}
\section{Introduction}\label{sec:intro}
\hyphenpenalty=500
\tolerance=5000
The target of handwritten mathematical expression recognition (HMER) is to generate markup sequences (e.g., LaTeX) from images containing handwritten mathematical expressions (HMEs).  Compared to traditional Optical Character Recognition (OCR) tasks, HMER suffers two difficulties: 1) The two-dimensional structure denotes the necessity to identify the layout patterns within HMEs. 2) Ambiguities and various writing styles will further cause confusion.

Grammar-based 
methods~\cite{alvaro2014recognition,maclean2013new,tang2024offline} attempt to solve these problems with grammatical structure analysis. However, these methods usually depend on specific syntactic design and lack consideration of various handwritten styles, leading to handwriting-unawareness. In recent years, encoder-decoder frameworks~\cite{deng2016you,li2022counting,zhang2018multi,zhao2021handwritten} have achieved great success on image-to-sequence tasks due to their data-driven capabilities and end-to-end benefits. These methods usually use an encoder to embed images as semantic vectors and an attention mechanism-based decoder to generate output markups. We argue that all previous methods primarily analyze HME in the spatial domain.

For HMEs, precisely capturing their two-dimensional structures is crucial for improving recognition accuracy. In this regard, frequency domain information, especially high-frequency components, possesses an inherent advantage. Regardless of the diversity in handwriting styles, frequency domain information enables the transformation of visual perspective into a representation of pixel variations, thus accurately capturing the contours of formulas.
 
In this paper, we provide a new perspective for HMER. We propose a plug-and-play method to \textbf{m}arry \textbf{f}requency domain with \textbf{H}MER, named MFH. By introducing frequency domain features for HMER, even without any specific design on grammar, existing frameworks can implicitly learn the layouts and logics of 2D formulas and improve recognition accuracy.

\begin{wrapfigure}{r}{5.5cm}
\vspace{-22pt} 
\centering
\includegraphics[scale=0.4]{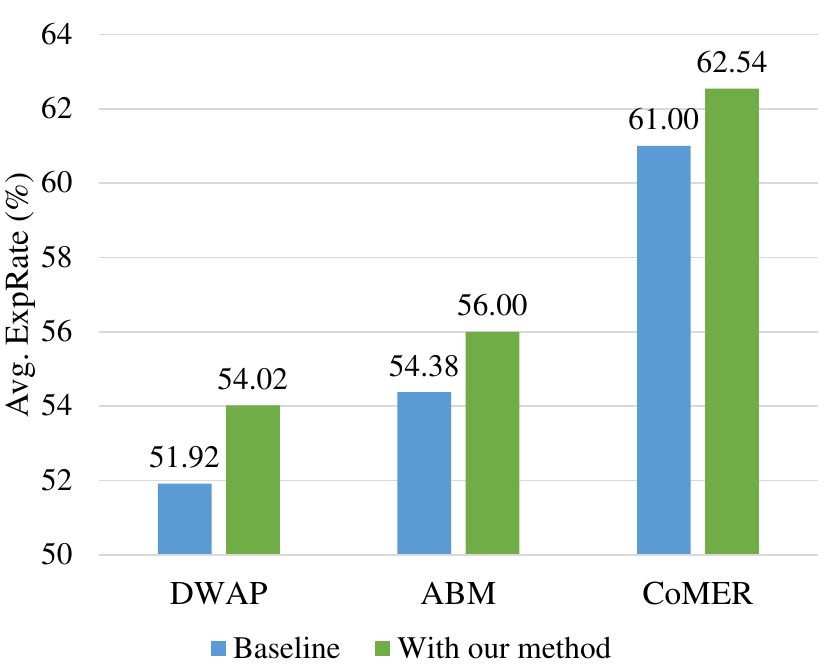}
\vspace{-16pt} 
\caption{Comparison of different baselines and our method.}
\vspace{-25pt} 
\label{fig:dct&idct}
\end{wrapfigure}
To demonstrate the effectiveness of our MFH, we conduct experiments on the CROHME datasets. As shown in Fig.~\ref{fig:dct&idct}, with state-of-the-art (SOTA) CoMER~\cite{zhao2022comer} as our baseline, MFH-CoMER achieves 61.66\%, 62.07\%, and 63.72\% on CROHME 2014/2016/2019, outperforming the baseline by 0.95\%, 2.09\%, and 1.50\%, respectively. Performances on DWAP~\cite{zhang2018multi} and ABM~\cite{bian2022handwritten} illustrate stable improvements of our method, which verifies its generalization on different baselines.

Our major contributions are summarized as follows: 1) We demonstrate the effectiveness of the frequency domain for HMER tasks and verify its potential for analyzing mathematical formulas. 2) We propose MFH, an approach that leverages frequency domain information and elevates the performance limit of HMER. MFH is plug-and-play and compatible with existing frameworks.



\section{Related Works}
\subsection{HMER Methods}
In the past few decades, many methods for HMER have been proposed. The early methods are mainly based on grammar. With the advancement of encoder-decoder architectures, methods based on this framework have emerged.
\subsubsection{Grammar-based Methods.} 
Grammar-based methods can be divided into three steps: symbol segmentation, symbol recognition, and grammar-based structural analysis. Kinds of hand-designed grammars, such as stochastic context-free grammar~\cite{alvaro2014recognition,yamamoto2006line}, relational grammar~\cite{maclean2013new}, graph grammar~\cite{lavirotte1998mathematical} and definite clause grammars~\cite{chan2000efficient,chan2001error} have been proposed. Although these methods based on predefined grammar show good interpretability, complex grammar rules and poor generalization limit their availability.
\subsubsection{Encoder-decoder-based Methods.} 
In recent years, the encoder-decoder-based methods~\cite{cheng2017focusing,yuan2020automatic,zhang2018multi} have gained satisfactory results in various image-to-sequence tasks. The encoder extracts features from an input image, and the decoder generates them as a sequence. Zhang \textit{et al.}~\cite{zhang2017watch} propose WAP with an FCN as the encoder and the coverage attention mechanism to alleviate the lack of coverage problem. Later, DenseWAP~\cite{zhang2018multi} utilizes DenseNet~\cite{huang2017densely} rather than VGG as the encoder in WAP and consequently improves the performance. Such DenseNet encoder design is adopted by many subsequent works~\cite{bian2022handwritten,chi2021handwritten,li2022counting}. With the utilization of Transformer~\cite{vaswani2017attention}, some methods~\cite{zhao2022comer,zhao2021handwritten} apply a transformer-based architecture instead of RNN~\cite{graves2013speech} as the decoder. The success of this paradigm also benefits from the introduction of larger datasets on various difficult tasks~\cite{dikubab2022comprehensive,kuang2023visual}. In this paper, however,we address the limitation of previous methods focusing solely on the spatial domain by introducing frequency domain analysis at the encoder side.

\subsection{Discrete Cosine Transform in Deep Learning} 
With the development of deep learning, discrete cosine transform (DCT) has been used in more and more computer vision tasks such as segmentation~\cite{shan2021decouple,shen2021dct} and deepfake detection~\cite{durall2020watch,giudice2021fighting}. CAT-Net~\cite{kwon2021cat} incorporates a DCT stream to learn compression artifacts based on binary volume representation of DCT coefficients, thus localizing spliced objects considering RGB and DCT domains jointly. In terms of document understanding, DocPedia~\cite{feng2023docpedia} opts to process visual input directly in the frequency domain rather than in pixel space. These approaches showcase DCT as a supplementary and alternative approach to traditional vision tasks. Inspired by their advancements, our method distinctively applies DCT to the frequency domain, which harnesses high-frequency information and elevates spatial structure analysis ability.

\section{Preliminaries}
\subsection{A Revisit of Discrete Cosine Transform}\label{sec_revisit}
Discrete cosine transform (DCT)~\cite{ahmed1974discrete} is a unique form of Fourier transform, which can operate on fixed-size patches for computing efficiency, named Patch-DCT. The transformed coefficients can be efficiently manipulated and encoded. Besides, DCT produces real-valued coefficients, simplifying the representation and storage of processed data.
With patch size $n$, Patch-DCT performs a transformation on each image patch, converting it from the spatial domain to the frequency domain:
 \begin{equation}
     F(u,v)= \frac{2}{n}C(u)C(v)\sum_{x=0}^{n-1}\sum_{y=0}^{n-1}f(x,y)cos[ \frac{(2x+1)u\pi}{2n}]cos[ \frac{(2y+1)v\pi}{2n}]  \label{eq:2d-dct}, 
      \end{equation}
\begin{equation}
    C(\alpha )\left\{\begin{matrix} 
  \frac{1}{\sqrt{2}} , \,\mathrm {if} \,\alpha =0 \\ 
  \ 1 ,\quad otherwise
\end{matrix}\right.,
\end{equation}
where $f(x, y)\in\mathbb{R}^{n\times n}$ is the input patch, and $F(u, v)\in\mathbb{R}^{n\times n}$ is the coefficient table representing the various frequency components.

In HMER, Patch-DCT is applied to non-overlapping patches of an image. The image is divided into square patches. Each patch is then transformed independently using Patch-DCT in Eq.~\ref{eq:2d-dct} to obtain frequency domain information.

\begin{figure}[t]
\begin{adjustbox}{center}
\centering
    \includegraphics[width=13cm]{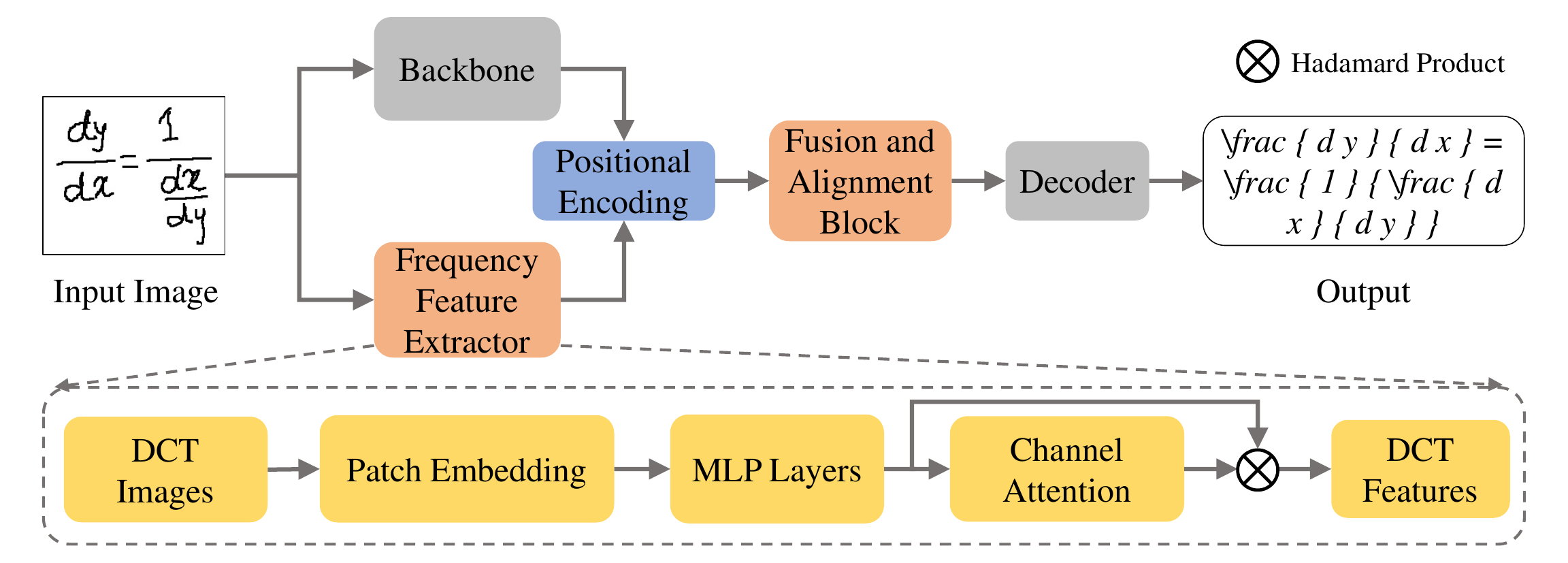}
\end{adjustbox}
\caption{The overview of MFH, which consists of a patch embedding layer, MLP layers,  channel attention mechanism, and a Fusion and Alignment Block (FAB).}
\label{fig:dct_extractor}
\end{figure}

\section{Our Method} \label{sec3}
The overview of our method is shown in Fig.~\ref{fig:dct_extractor}. We first obtain high-frequency-retained patches by Patch-DCT and a simple selection strategy of coefficients (Sec.~\ref{sec:data_proc}). Then, the patches go through the extractor pipeline (detailed in Sec.~\ref{sec:extractor}) to obtain the frequency features. We elaborate a Fusion and Alignment Block (FAB) in Sec.~\ref{sec:FAB} to align features of two domains. Our MFH can be easily plugged into mainstream HMER frameworks~\cite{bian2022handwritten,zhang2018multi,zhao2022comer}.
\vspace{-4pt}
\subsection{Transformation towards Frequency Domain}\label{sec:data_proc}
\begin{figure}[t]
\begin{adjustbox}{center}
    \centering
\includegraphics[width=13cm]{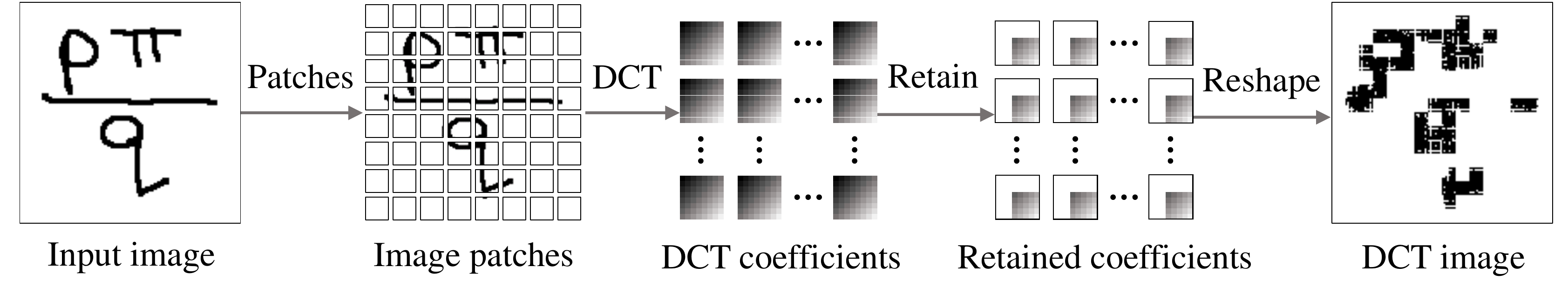}
\end{adjustbox}    
    \caption{The DCT pre-processing. The input image is first divided into patches, and DCT is applied to each patch. Subsequently, high-frequency components are retained, and low-frequency ones are set to zero. Finally, patches are reshaped to their original size. }
    \label{fig:data processing}
\end{figure}

In our pre-processing shown in Fig.~\ref{fig:data processing}, the input image $I\in\mathbb{R}^{1\times H\times W}$ is first divided into $\frac{H}{n}\times \frac{W}{n}$ patches for Patch-DCT application (Sec.~\ref{sec_revisit}), where $n$ is the patch size. For each patch, we implement DCT with Eq.~\ref{eq:2d-dct}, obtaining a $n\times n$ DCT coefficient table. Each DCT coefficient is a sum of cosine functions oscillating at a specific frequency, representing the amplitude of a frequency component in this patch.
As demonstrated in Sec.~\ref{sec:intro}, high-frequency information helps capture the contours of formulas, so we focus on the preservation of high-frequency components.

Specifically, for a $n\times n$ DCT coefficient table, low-frequency coefficients are positioned in the top-left quadrant, representing the overall changes and flat parts of the image. High-frequency coefficients, on the contrary, are situated in the bottom-right quadrant within an image for details and areas of significant variation. So we select coefficients in the bottom-right $m \times m$ $(m \le n)$ area by setting coefficients outside this area to zero.

Finally, we get the high-frequency-retained patches. They are then combined into their original dimensional form, denoted as $I^{'} \in \mathbb{R}^{1 \times H \times W}$.

\subsection{Frequency Feature Extractor}\label{sec:extractor}
A frequency feature extractor is proposed as our pipeline for the processed feature after Patch-DCT. As illustrated in Fig.~\ref{fig:dct_extractor}, our frequency feature extractor consists of a patch embedding layer, multiple MLP (Multi-Layer Perceptron) layers, a channel attention layer, and a positional encoding layer. 

\subsubsection{Patch Embedding.}
We first tokenize each $n\times n$ patch with a patch-embedding layer, aggregating local interaction information into the channel dimension. Specifically, it is a convolutional layer with kernel size $n$ and stride $n$, extracting ${I^{'}}$ as $\mathcal{P} \in \mathbb{R}^{C \times \frac{H}{n} \times\frac{W}{n}}$. Each $n\times n$ patch is embedded as a token.
\vspace{-8pt}
\subsubsection{MLP Layers.}
For tokens $\mathcal{P}$, we choose simple MLP layers to encode frequency domain information in the channel dimension. Tokens $\mathcal{P}$ are first processed by a layer normalization. After this are MLP layers containing fully connected layers, activation functions, and random dropouts. Finally, we get the output feature $\mathcal{P^{’} }$ with a residual link.
\vspace{-8pt}
\subsubsection{Channel Attention.}
With feature $\mathcal{P}'$, the channel-wise attention is adopted to enhance and select specific channel properties. We get the enhanced feature $K$ as follows:
\begin{equation}
    \begin{array}{c}
        \mathbf{\upsilon} = \mathcal{R} (W_{1}(\mathcal{G}, (\mathbf{\mathcal{P}'}))),\\
        K= \mathbf{\mathcal{P}'} \otimes \mathcal{S}(W_{2}(\mathbf{\upsilon})),
    \end{array}
\end{equation}
where $\mathcal{G}$ is the global average pooling. $\mathcal{R}$ and $\mathcal{S}$ refer to ReLU activation and Sigmoid function, respectively. $\otimes$ means Hadamard Product. $W_{1}$ and $W_{2}$ are both trainable weights.
\vspace{-8pt}
\subsubsection{Positional Encoding.}
Before decoding, we use the same positional encoding method as~\cite{carion2020end,zhao2022comer,zhao2021handwritten} to encode positional information of the input feature. Given position $x$ and index $i$ of feature dimension, 1D positional encoding with dimension size $d$ is defined as:
\begin{equation}
    \begin{array}{c}
        {p}_{x} [2i]=sin(p/10000^{2i/d}),\\
        {p}_{x} [2i+1]=cos(p/10000^{2i/d}),
    \end{array}
\end{equation}
where we get the positional encoding vector $p_{x}$. For 2D feature $K$, a relative positional encoding is applied by a combination of two 1D positional encodings:
\begin{equation}
\begin{array}{c}
    \bar{x} =\frac{x}{h}, \bar{y} =\frac{y}{w},\\
{p}_{(x,y)} =[{p}_{\bar{x},d/2},  {p}_{\bar{y},d/2}],
    \end{array}
\end{equation}
where tuple $(x,y)$ is the normalized 2D coordinates. $\bar{x}$ and $\bar{y}$ represent the relative position to output feature. $[, ]$ is the concatenation operation. 
 
\begin{figure}[t]
\begin{adjustbox}{center}
    \centering
    \includegraphics[height=5.5cm]{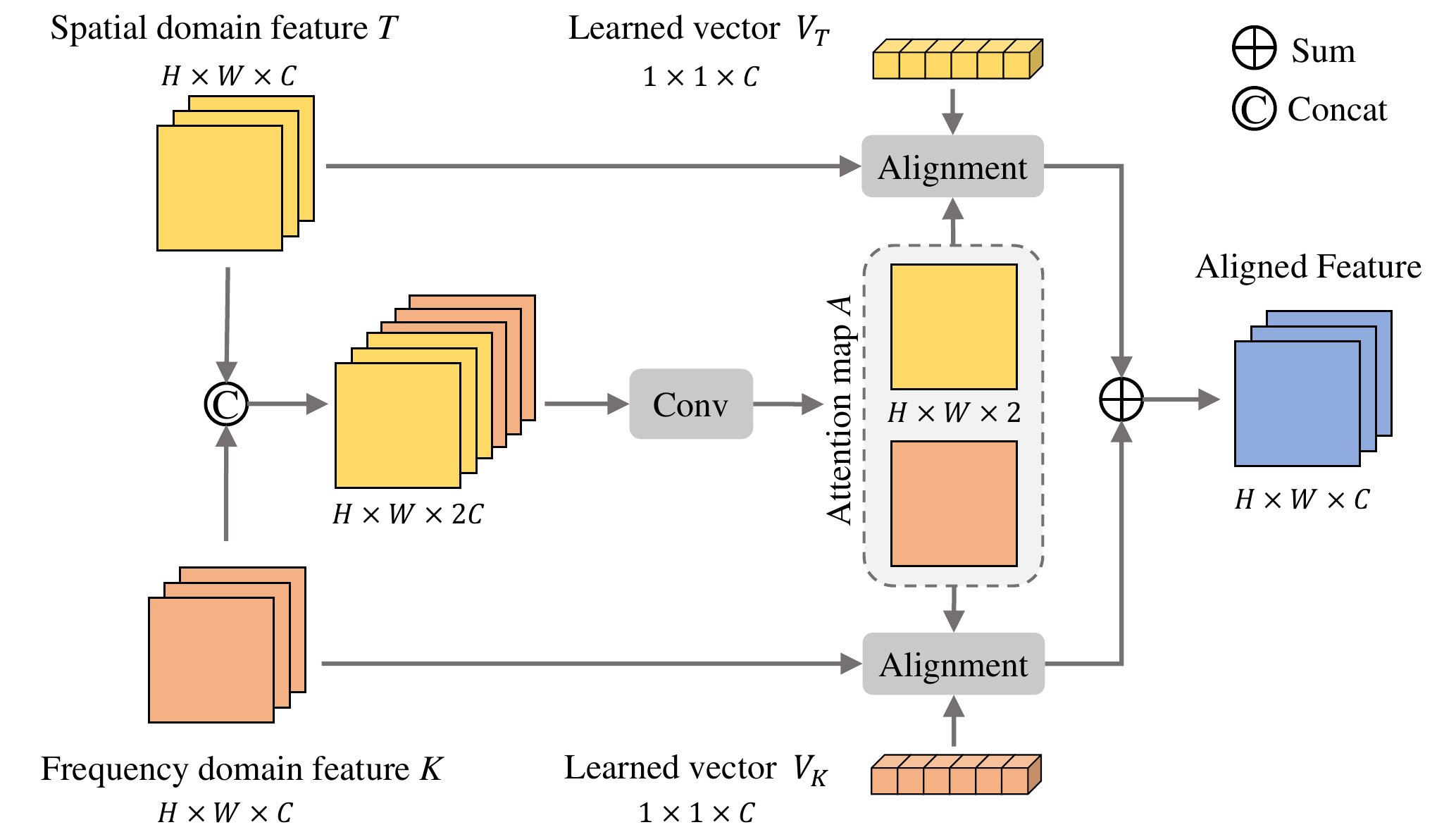}    
\end{adjustbox}
    \caption{The FAB used in the feature fusion stage.}
    \label{fig:FAB}
\end{figure}

\subsection{Fusion and Alignment Block}\label{sec:FAB}
Considering that most of the current HMER models adopt a generic CNN-based network as the backbone (e.g., DenseNet~\cite{huang2017densely} in DWAP~\cite{zhang2018multi}, CAN~\cite{li2022counting} and CoMER~\cite{zhao2022comer}), it is logical to implement our method (as frequency domain stream) in parallel with this kind of encoder (as spatial domain stream). 

For Patch-DCT, we implement a padding operation on the input image as the input size may not be divisible by patch size $n$, introducing a minor alteration in the data. However, the same image will pass through the CNN backbone unchanged. Thus, misalignment will exist between feature $T$ outputted by the CNN backbone and the frequency domain feature $K$, even if they share the same size $K, T\in \mathbb{R}^{C \times \frac{H}{16} \times \frac{W}{16}}$ after $16\times$ downsampling.

In this part, we specify a Fusion and Alignment Block (FAB) to implement alignment. We argue that the $n\times n$ divided treatment makes $K$ focus on localized properties and can compensate local texture information for $T$. As shown in Fig.~\ref{fig:FAB}, our FAB outputs a two-channel attention map ${A}$ via a downsampling convolution layer. It fuses and compresses the concatenation of $T$ and $K$. Then, we implement alignment by reinforcing each channel to pay attention only to one single domain:
\begin{equation}
\begin{array}{c}
K=A_{1} K\otimes V_{K},\\  
T=A_{2} T\otimes V_{T}, \label{eq:fusion}
\end{array}
\end{equation}
where $A_{1}$, $A_{2}$ refer to the first and second channel of attention map ${A}$, respectively. $V_{K}$, $V_{T}$ are trainable attention values for each channel.

Finally, we add $K$ and $T$ directly since trainable weights $V_{K}$ and $V_{T}$ in Eq.~\ref{eq:fusion} have linearly blended features of two domains. The mentioned alignment operations cause no damage to the end-to-end property as we merely refine the output of the encoder side with frequency domain information.

\section{Experiment}
\subsection{Implementation Details} 
In Patch-DCT, we set patch size $n$ to 8 and high-frequency retention number $m$ to 5. In the frequency feature extractor, the output channel dimension of patch embedding layer $C$ is set to 256, and the dropout rate in MLP layers to 0.3. In MFH-CoMER, we use 6$\times$ MLP layers. To validate the generalization and portability of our method, we migrate our method to DWAP~\cite{zhang2018multi} and ABM~\cite{bian2022handwritten} with minor parameter adjustments. Experiments are conducted on four NVIDIA RTX 3090 GPUs with 24 GB memory.

\subsection{Datasets and Evaluation Metrics}
The Competition on Recognition of Online Handwritten Mathematical Expressions (CROHME) datasets are open datasets for HMER tasks. The training data contains 8836 binary images, and the test sets 2014/2016/2019 contain 986/1147/1199 binary images.

As for evaluation metrics, we choose “ExpRate”, “$\le$ 1 error” and “$\le$ 2 error” to measure the performance of the proposed method, indicating we tolerate 0 to 2 symbol-level errors, respectively. The “$\le$ 1 error” and “$\le$ 2 error” are calculated with dynamic programming. We use CROHME 2014 as the validation set to select the best-performing model during training.

\begin{table}[t]
\renewcommand{\arraystretch}{1.3}
\setlength{\tabcolsep}{0.3mm}
\caption{Results on CROHME datasets. † means our reproduced results. MFH-DWAP and MFH-ABM represent using DWAP~\cite{zhang2018multi} and ABM~\cite{bian2022handwritten} as baselines, respectively. MFH-CoMER follows the same data augmentation in CoMER~\cite{zhao2022comer}.}

\centering
\label{tab2} 
\begin{adjustbox}{center}
\scalebox{0.76}{
    \begin{tabular}{l c c c c c c c c c c}  
    \Xhline{1.2pt}
    \multirow{2}*{Method}&\multirow{2}*{Year}&\multicolumn{3}{c}{CROHME 2014}&\multicolumn{3}{c}{CROHME 2016}&\multicolumn{3}{c}{CROHME 2019}\\   
    \cmidrule{3-11}
    & &ExpRate$\uparrow$& \ $\le\ $1$\ \uparrow$ \ & \ $\le\ $2$\ \uparrow$ \ 
    &ExpRate$\uparrow$& \ $\le\ $1$\ \uparrow$ \ & \ $\le\ $2$\ \uparrow$ \ 
    &ExpRate$\uparrow$& \ $\le\ $1$\ \uparrow$ \ & \ $\le\ $2$\ \uparrow$ \ \\
    \midrule
    \rowcolor{gray!20}\multicolumn{11}{l}{Without data augmentation}\\ 
    DWAP~\cite{zhang2018multi}&ICPR 18&50.10&-&-&47.50&-&-&-&-&-\\
    BTTR~\cite{zhao2021handwritten}&ICDAR 20&53.96&66.02&70.28&52.31&63.90&68.61&52.96&65.97&69.14\\
    TSDNet~\cite{zhong2022tree}&ACM 22&54.70&68.85&74.48&52.48&68.26& 73.41&56.34&72.97&77.84\\
    SAN~\cite{yuan2022syntax}&CVPR 22&56.20& 72.60&79.20&53.60& 69.60&76.80&53.50&69.30&70.10\\
    ABM~\cite{bian2022handwritten}&AAAI 22&56.85&73.73&81.24&52.92&69.66&78.73&53.96&71.06&78.65\\
    CAN-ABM~\cite{li2022counting}&ECCV 
    22&57.26&74.52&82.03&56.15&72.71&80.30&55.96&72.73&80.57\\
    Liu et al.~\cite{liu2022semantic}&PRCV 22&53.91&-&-&52.75&-&-&-&-&-\\
    Han et al.~\cite{han2022handwritten}&PRCV 22&56.80&71.27&76.85&53.34&67.56&74.19&54.62&68.97&74.64\\
    SAM-CAN~\cite{liu2023semantic}&ICDAR 23 &58.01&-&-&56.67&-&-&57.96&-&-\\
    GETD~\cite{tang2024offline}&PR 24&53.45&67.54& 72.01&55.27&68.43&72.62&54.13&67.72&71.81\\
    BDP~\cite{li2024tree}&PR 24&57.71& 73.53&80.83&55.62& 71.67&78.73&59.47&75.23&80.90\\
    \midrule
    DWAP(baseline)$^{\dagger}$&ICPR 18&51.12&64.91&74.54&52.57&65.91&74.37&51.96&66.97&75.23\\
    \textbf{MFH-DWAP(ours)}&-&53.25&67.24&75.46&54.66&68.35&75.85&54.05&68.14&76.31\\
    \midrule
    ABM(baseline)$^{\dagger}$&AAAI 22&55.88&72.72&80.32&52.40&70.62&78.20&55.05&74.98&80.65\\
    \textbf{MFH-ABM(ours)}&-&57.00&73.02&80.53&55.10&72.10&80.38&56.05&74.40&81.99\\
    \midrule
    \rowcolor{gray!20}\multicolumn{11}{l}{With data augmentation}\\ 
    Li et al.~\cite{li2020improving}&ICFHR 20&56.59&69.07&75.25&54.58&69.31&73.76&-&-&-\\
    Ding et al.~\cite{ding2021encoder}&ICDAR 21&58.72&-&-&57.72&70.01&76.37& 61.38&75.15&80.23\\
    CoMER~\cite{zhao2022comer}&ECCV 22&59.33&71.70&75.66&59.81&74.37&80.30&62.97&77.40&81.40\\
    \midrule
    CoMER(baseline)$^{\dagger}$&ECCV 22&60.71&76.35&83.35&59.98&76.63&83.44&62.22&79.98&85.15\\
    \textbf{MFH-CoMER(ours)}&-&\textbf{61.66}&\textbf{76.88}&\textbf{83.37}&\textbf{62.07}&\textbf{78.29}&\textbf{84.92}&\textbf{63.72}&\textbf{81.40}&\textbf{86.74}\\
    \Xhline{1.2pt}
    \end{tabular}
}
\end{adjustbox}
\end{table}

\subsection{Comparison with Existing Methods}
In this section, we compare our method with existing models. As MFH is plug-and-play, we can easily insert it into different frameworks~\cite{bian2022handwritten,zhang2018multi,zhao2022comer}. To keep fair, we divide previous methods into two categories based on whether they implement data augmentation. The results are shown in Tab. \ref{tab2}.

Experiments show that our MFH-DWAP and MFH-ABM outperform their baselines~\cite{bian2022handwritten,zhang2018multi}. This verifies our method's generalization and stable improvement based on existing frameworks. We further conduct experiments on the latest SOTA method CoMER~\cite{zhao2022comer}, which adopts a data augmentation. We find that MFH-CoMER outperforms its baseline by 0.95\%, 2.09\%, and 1.50\% on three test sets, respectively, emphasizing that our method can also be migrated to a data augmentation strategy.
 
\subsection{Analysis}
We conduct ablation studies and component analysis on CROHME datasets with MFH-CoMER. Notably, we select the average expression recognition accuracy (Avg. ExpRate) on three test sets as the evaluation metric.

\subsubsection{Ablation of Frequency Domain Information.}
We first verify whether the introduction of frequency domain information is the principal factor in improving performance. Specifically, we  substitute DCT-processed images with original input in Fig. \ref{fig:dct_extractor}. The results are detailed in Tab. \ref{tab:dct images}. We can confirm that the extra parameters by our MFH merely brings a slight improvement. It is the integration of frequency-domain information, not the parameters increment, that significantly boosts the network’s capability. 

\subsubsection{Ablation of Different 
Frequency Domain Transformations.}
In this paper, we choose DCT as the bridge to gain frequency domain information during pre-processing. To demonstrate the effectiveness of DCT, we compare it with fast Fourier transform (FFT). The results are shown in Tab.~\ref{tab:fft}, which proves that DCT is more suitable for our MFH than FFT in image processing. We presume that DCT does not involve any complex number operation, which matches real-valued math expression images better than FFT.

\subsubsection{Ablation of Different Patch Sizes.}
We divide input images into $8\times 8$ patches in Sec.~\ref{sec:data_proc}. To explore the impact of different patch sizes, we conduct experiments with patch sizes $8$ and $16$. Results are shown in Tab. \ref{tab:ablation5}, which show that $8\times 8$ gains higher average ExpRate. 
\begin{table}[!t]
\scriptsize
\setlength{\tabcolsep}{1mm}
\begin{minipage}{0.38\linewidth}
\renewcommand{\arraystretch}{1.35}
\caption{Ablation study of frequency-domain information in MFH.}
\begin{adjustbox}{center}
\centering
\scalebox{1.05}{
    \begin{tabular}{l c}
    \Xhline{1.2pt}
    {Method}&ExpRate$\uparrow$\\  
    \midrule
    CoMER~\cite{zhao2022comer} (baseline)&61.00 \\
    $+$ Original images&61.32 \\
    $+$ DCT images (ours)&\textbf{62.54} \\
    \Xhline{1.2pt}
    \end{tabular}
}
\end{adjustbox}
\label{tab:dct images}

\end{minipage}
\hfill
\begin{minipage}{0.3\linewidth}
\renewcommand{\arraystretch}{1.8}
\caption{Comparison of two frequency transformations.}
\scriptsize
\begin{adjustbox}{center}
\setlength{\tabcolsep}{1mm}
\centering
\scalebox{1.05}{
    \begin{tabular}{c c c}
    \Xhline{1.2pt}
    DCT&FFT&ExpRate$\uparrow$\\
    \midrule
    \textbf{-}&\Checkmark&61.73 \\
    \Checkmark&\textbf{-}&\textbf{62.54} \\
    \Xhline{1.2pt}
    \end{tabular}
    }
\end{adjustbox}
\label{tab:fft}
\end{minipage}
\hfill
\begin{minipage}{0.28\linewidth}
\renewcommand{\arraystretch}{1.8}
\caption{Comparison of different patch sizes in Patch-DCT.}
\begin{adjustbox}{center}
\centering
\scalebox{1.05}{
    \begin{tabular}{c c}
    \Xhline{1.2pt}
    {Patch Size}&ExpRate$\uparrow$\\  
    \midrule
    $8\times 8$&\textbf{62.54}\\
    $16\times 16$&61.22\\
    \Xhline{1.2pt}
    \end{tabular}
    }
\end{adjustbox}
\label{tab:ablation5}
\end{minipage}
\end{table}

\subsubsection{Effects of Frequency Domain Information Retention.} \label{sample_num}
\begin{figure}[t]
\begin{adjustbox}{center}
\includegraphics[width=10cm]{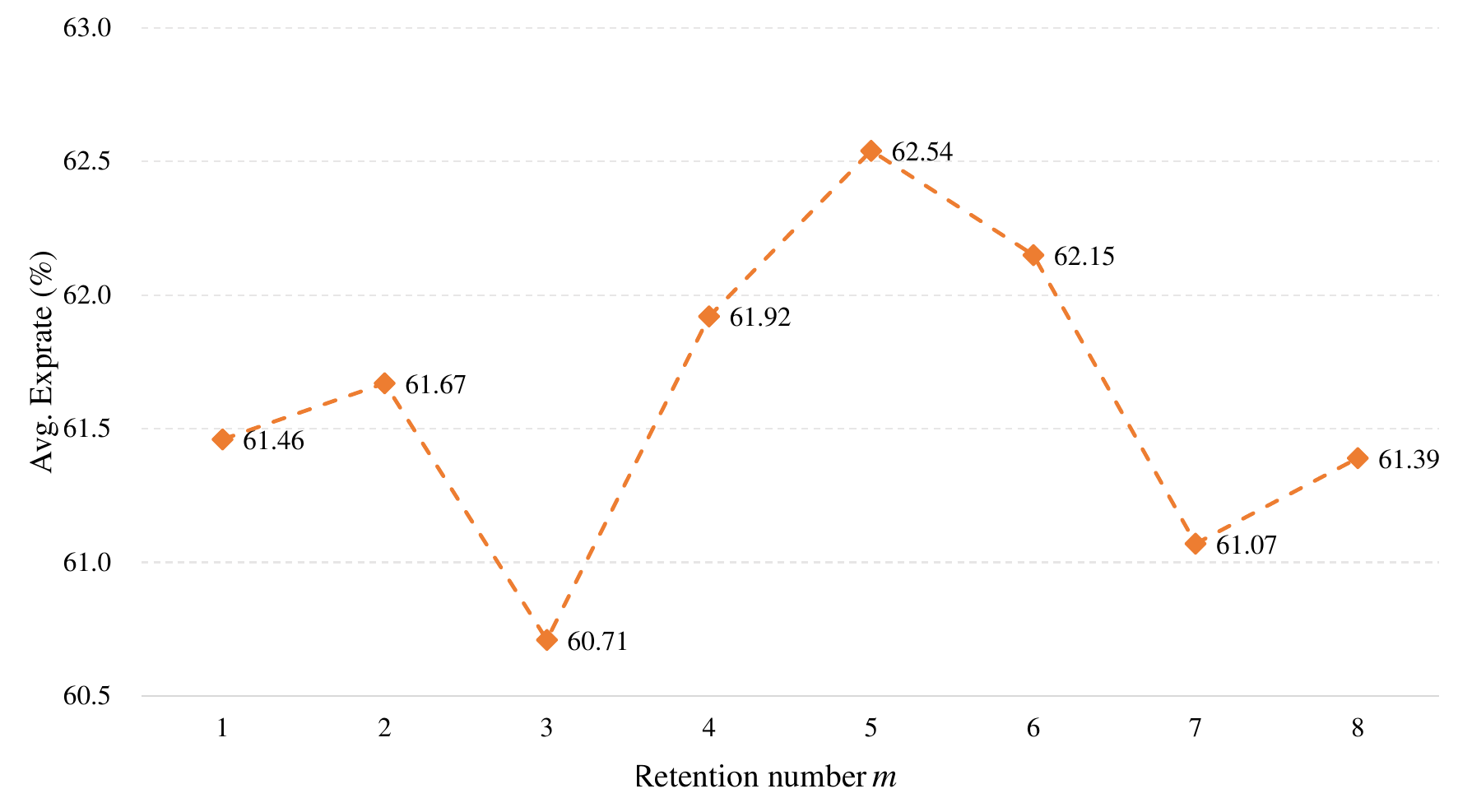}
\end{adjustbox}
\caption{Avg. ExpRate for different retention number $m$.}
\label{fig:retention}
\end{figure}

In data pre-processing, we decide how much high-frequency information to retain in the transformed patches by setting the integer retention number $m$. In this section, we conduct experiments on MFH-CoMER using different values of $m$, which means retaining $m\times m$ coefficients in the right-down corner of the patches. As shown in Fig.~\ref{fig:retention}, setting $m$ to 5 yields the highest Avg. ExpRate. Notably, there is no frequency information to be discarded if $m$ is set to $n$. The non-monotonic curve suggests that an optimal point must be found for high-frequency retention. 

\subsubsection{Analysis of Different Components.} In our MFH, channel attention selects and enhances specific channel properties. Positional encoding compensates for the lack of relative position understanding attributed to tokenization processing. We also demonstrate in Sec.~\ref{sec:FAB} that the proposed Fusion and Alignment Block (FAB) ensures features of two domains are mutually informative. 
In this part, we conduct experiments to validate the effectiveness of these components. Results in Tab.~\ref{tab:ablation1} show that both channel attention and positional encoding consistently improve recognition accuracy. With the implementation of FAB for feature alignment, the Avg. ExpRate achieves a peak of $62.54\%$. 

\subsubsection{Analysis of Alignment Method in FAB.}
In Sec.~\ref{sec:FAB}, we propose Fusion and Alignment Block (FAB) to align features of two domains. As shown in Tab. \ref{tab:ablation2}, implementing concatenation rather than direct addition and utilizing learnable vectors both obtain better results. We infer that utilizing concatenation before obtaining attention map ${A}$ is a better choice, as adding the features directly will cause a loss of bimodal information. Experimental results reveal that learnable vectors balance the trade-offs concerning information usage between two domains, substantially outperforming their non-learnable counterparts. This indicates that our FAB has a well-designed construction to realize alignment.
\begin{table}[t]
\scriptsize
\setlength{\tabcolsep}{1mm}
\begin{minipage}{0.5\linewidth}
\renewcommand{\arraystretch}{1.25}
\caption{Analysis of different components. The first line refers to the baseline CoMER~\cite{zhao2022comer}. MFH is our method. }
\begin{adjustbox}{center}
\centering
\begin{tabular}{c | c c c c}
\Xhline{1.2pt}
{MFH}&{Channel-Att}&{Pos-Enc}&{FAB}&Avg. ExpRate$\uparrow$\\  
\hline
\textbf{-}&\textbf{-}&\textbf{-}&\textbf{-}&61.00(baseline)\\
\Checkmark&\textbf{-}&\textbf{-}&\textbf{-}&61.52\\
\Checkmark&\Checkmark&\textbf{-}&\textbf{-}&61.73\\
\Checkmark&\textbf{-}&\Checkmark&\textbf{-}&61.77\\
\Checkmark&\Checkmark&\Checkmark&\textbf{-}&62.04 \\
\Checkmark&\Checkmark&\Checkmark&\Checkmark&\textbf{62.54} \\
\Xhline{1.2pt}
\end{tabular}
\end{adjustbox}
\label{tab:ablation1}
\end{minipage}
\hfill
\begin{minipage}{0.42\linewidth}
\renewcommand{\arraystretch}{1.7}
\setlength{\tabcolsep}{0.8mm}
\caption{Analysis of alignment method in FAB. Concat and Vectors refer to concatenation and learnable vectors, respectively.}
\begin{adjustbox}{center}
\centering
\begin{tabular}{c c c}
\Xhline{1.2pt}
{Concat}&{Vectors}&ExpRate$\uparrow$\\  
\midrule
\Checkmark&\textbf{-}&61.23 \\
\textbf{-}&\Checkmark&62.36 \\
\Checkmark&\Checkmark&\textbf{62.54} \\
\Xhline{1.2pt}
\end{tabular}
\end{adjustbox}
\label{tab:ablation2}
\end{minipage}
\end{table}

\begin{figure}[t]
\begin{adjustbox}{center}
\includegraphics[height=5.5cm]{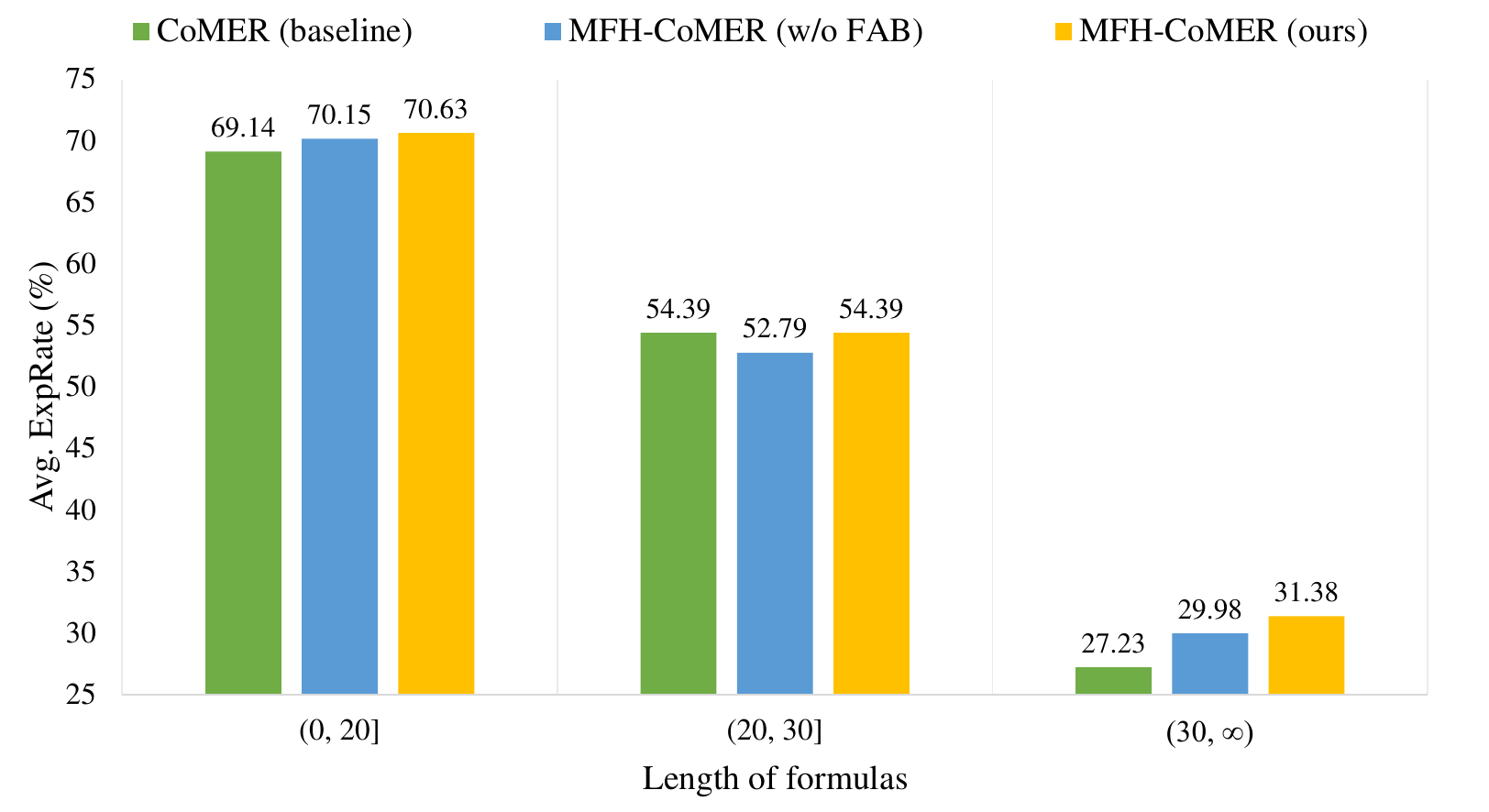}
\end{adjustbox}
\caption{Avg. ExpRate for different lengths of formulas.}
\label{fig:length_error}
\end{figure}

\begin{figure}[h]
\begin{adjustbox}{center}
\includegraphics[width=10cm]{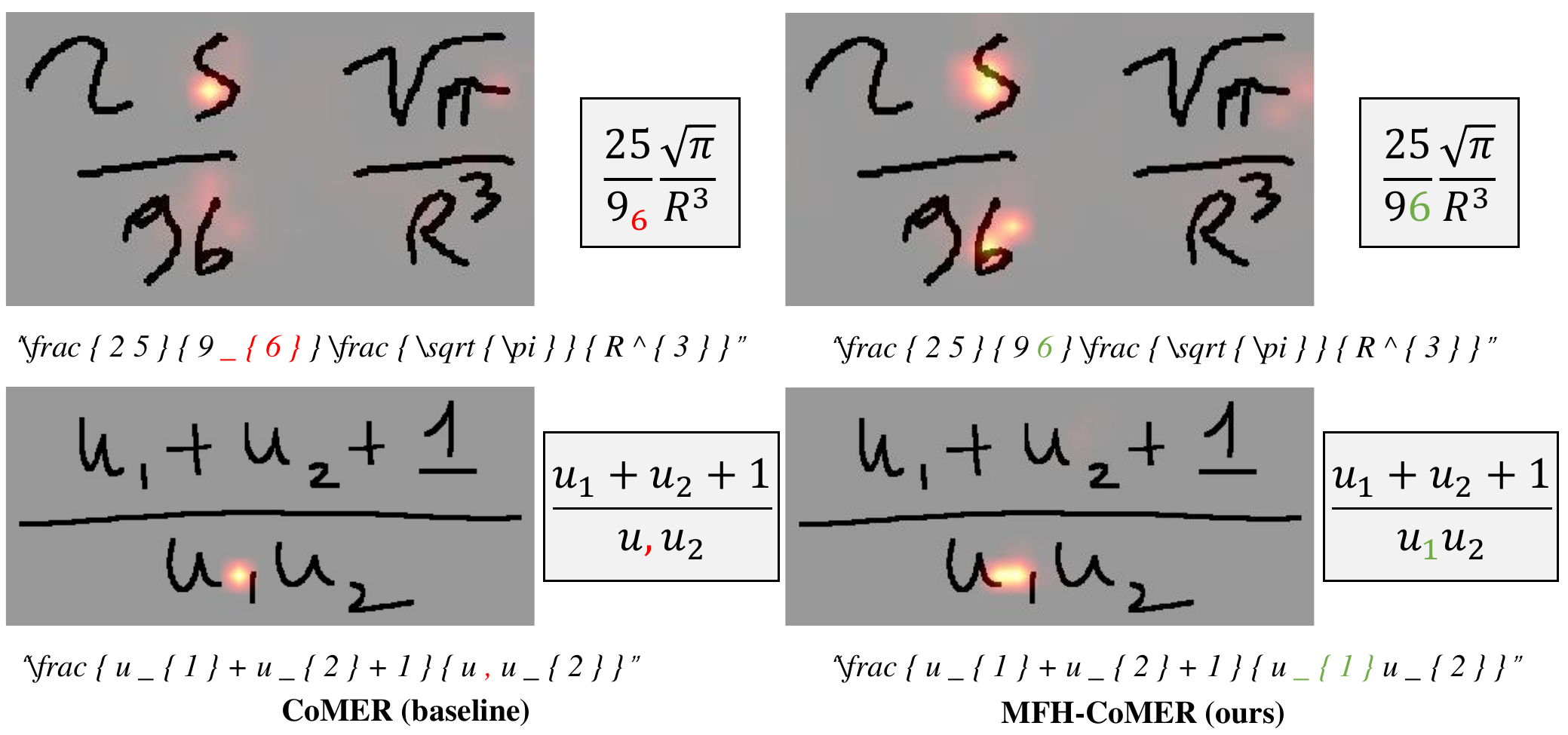}
\end{adjustbox}
\caption{Attention maps of CoMER~\cite{zhao2022comer} and our MFH-CoMER.}
\label{fig:error_example}
\end{figure}

\subsection{Performance for Different Lengths of Formulas}
We hypothesis in Sec.~\ref{sec:intro} that frequency information implicitly learns the layout and logic of 2D formulas. This suggests that our MFH is poised to improve recognition accuracy for long sequences, typically characterized by more complex spatial structures. The results are shown in Fig.~\ref{fig:length_error}. We divide formulas in test sets into three intervals, which represent the short, medium, and long lengths of formulas, respectively. MFH-CoMER outperforms the baseline on formulas with long length $(\ge 30)$ by a large margin. This is attributed to the fusion of spatial domain and frequency domain features. Specifically, we visualize attention maps with two example in Fig.~\ref{fig:error_example}, where the red symbols indicate errors caused by baseline model CoMER~\cite{zhao2022comer} and the green are corrections with our MFH-CoMER. For the example above, our method pays more precise attention to the symbol “6”. The baseline misses the accurate location for symbol “6” and mistakenly decodes it as a subscript of symbol “9”. For the example below, our method precisely identified “1” as the subscript of “u”, whereas the baseline incorrectly recognizes the subscript '1' as a comma. 

\subsection{Limitations}
Although introducing frequency-domain information for HMER improves model performance, further exploitation remains untapped. We have not explored the combination of frequency domain and attention-based mechanism on the decoder side. Besides, our method favors simplicity while sacrificing the perception of details. Perhaps a more elaborate design would enable precise correspondence between frequency components and specific symbols. 

\section{Conclusion} 
In this paper, we propose a Plug-and-Play method named MFH to utilize frequency domain information for HMER. By introducing frequency information to facilitate model training, our MFH improves the performance of different baselines stably. Experiments on the benchmark dataset CROHME validate the effectiveness of our method. We hope that frequency domain analysis can inspire subsequent work on HMER.

\subsubsection{Acknowledgements.}This work was supported by the NSFC (Grant No.623B2038) and in part by the Taihu Lake Innovation Fund for Future Technology (HUST: 2023-A-1).


\bibliographystyle{splncs04}
\end{document}